\documentclass[runningheads]{llncs}
\usepackage{graphicx}
\usepackage{amsmath,amssymb} 
\usepackage{multirow}
\usepackage{wrapfig}
\usepackage{color}
\usepackage[outercaption]{sidecap}
\usepackage[width=122mm,left=12mm,paperwidth=146mm,height=193mm,top=12mm,paperheight=217mm]{geometry}
\usepackage{makecell}

\begin{document}
\pagestyle{headings}
\mainmatter

\title{Zero-Shot Object Detection: Learning to Simultaneously Recognize and Localize  Novel Concepts} 

\titlerunning{Zero-Shot Object Detection}

\authorrunning{Shafin Rahman, Salman Khan and Fatih Porikli}

\author{Shafin Rahman$^{\dagger\star}$, Salman Khan$^{\star\dagger}$ and Fatih Porikli$^{\dagger}$}


\institute{$^{\dagger}$Australian National University\\
	$^{\star}$DATA61, CSIRO
}

\maketitle

\begin{abstract}
Current Zero-Shot Learning (ZSL) approaches are restricted to recognition of a single dominant unseen object category in a test image. We hypothesize that this setting is ill-suited for real-world applications where unseen objects appear only as a part of a complex scene, warranting both the `recognition' and `localization' of an unseen category. To address this limitation, we introduce a new \emph{`Zero-Shot Detection'} (ZSD) problem setting, which aims at simultaneously recognizing and locating object instances belonging to novel categories without any training examples. We also propose a new experimental protocol for ZSD based on the highly challenging ILSVRC dataset, adhering to practical issues, e.g., the rarity of unseen objects. To the best of our knowledge, this is the first end-to-end deep network for ZSD that jointly models the interplay between visual and semantic domain information. To overcome the noise in the automatically derived semantic descriptions, we utilize the concept of meta-classes to design an original loss function that achieves synergy between max-margin class separation and semantic space clustering. Furthermore, we present a baseline approach extended from recognition to detection setting. Our extensive experiments show significant performance boost over the baseline on the imperative yet difficult ZSD problem.
\keywords{Zero-shot learning, Object detection, Zero-shot detection}
\end{abstract}

\section{Introduction} \vspace{-0.5em}
Since its inception, zero-shot learning research has been dominated by the object classification problem \cite{Akata_PAMI_2016,Changpinyo_2016_CVPR,DeViSE_NIPS_2013,Kodirov_2015_ICCV,Lampert_PAMI_2014,Ba_CVPR_2015,bucher_ECCV_2016,norouzi_arXiv_2013,romera_ICML_2015,Xian_CVPR_2017,Zhang_2017_CVPR,Zhang_2015_ICCV,Zhang_2016_CVPR}. Although it still remains as a challenging task, the zero-shot recognition has a number of limitations that render it unusable in real-life scenarios. \emph{First}, it is destined to work for simpler cases where only a single dominant object is present in an image. \emph{Second}, the attributes and semantic descriptions are relevant to individual objects instead of the entire scene composition. \emph{Third}, zero-shot recognition provides an answer to unseen categories in elementary tasks, e.g., classification and retrieval, yet it is unable to scale to advanced tasks such as scene interpretation and contextual modeling, which require a fundamental reasoning about all salient objects in the scene. \emph{Fourth}, global attributes are more susceptible to background variations, viewpoint, appearance and scale changes and practical factors such as occlusions and clutter. As a result, image-level ZSL fails for the case of complex scenes where a diverse set of competing attributes that do not belong to a single image-level category would exist. 

\begin{figure}[t]
  \begin{center}
   \includegraphics[width=1\linewidth,trim={0cm 0cm 0cm 0cm},clip]{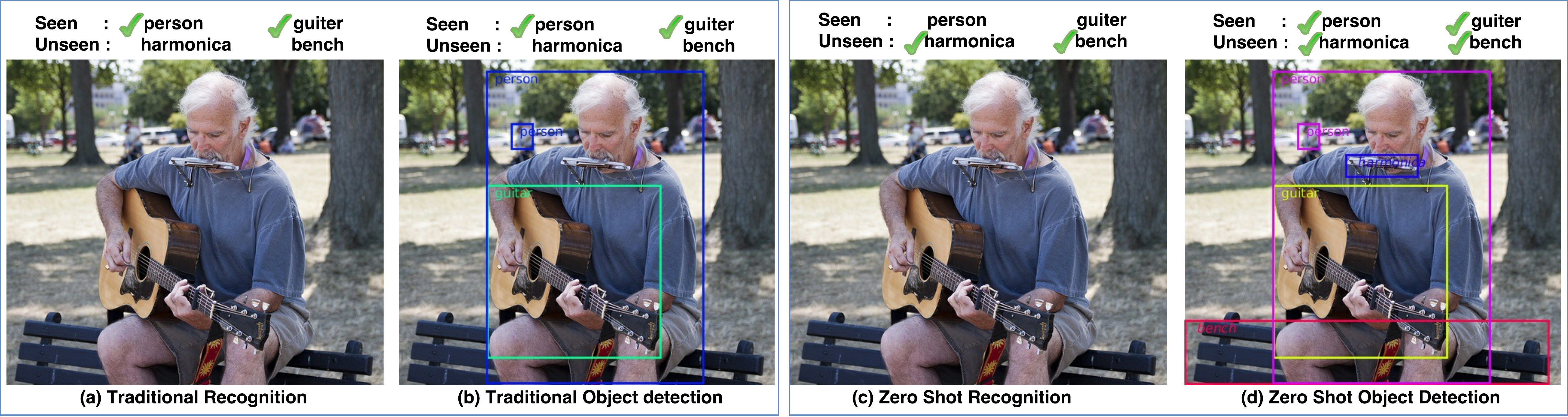}
  \end{center}
  \vspace{-2em}
   \caption{ZSD deals with a more complex label space (object labels and locations) with considerably less supervision (i.e., no examples of unseen classes). \textbf{(a)} Traditional recognition task only predicts seen class labels. \textbf{(b)} Traditional detection task predicts both seen class labels and bounding boxes. \textbf{(c)} Traditional zero-shot recognition task only predicts unseen class labels. \textbf{(d)} The proposed ZSD predicts both seen and unseen classes and their bounding boxes. }
\label{fig:motivation}
\end{figure}

To address these challenges, we introduce a new problem setting called the \emph{zero-shot object detection}. As illustrated in Fig.~\ref{fig:motivation}, instead of merely classifying images, our goal is to simultaneously detect and localize each individual instance of new object classes, even in the absence of any visual examples of those classes during the training phase. In this regard, we propose a new zero-shot detection protocol built on top of the ILSVRC - Object Detection Challenge \cite{ILSVRC_2015}. The resulting dataset is very demanding due to its large scale, diversity, and unconstrained nature, and also unique due to its leveraging on WordNet semantic hierarchy \cite{Wordnet_1995}. Taking advantage of semantic relationships between object classes, we use the concept of `\emph{meta-classes}'\footnote{Meta-classes are obtained by clustering semantically similar classes.} and introduce a novel approach to update the semantic embeddings automatically. Raw semantic embeddings are learned in an unsupervised manner using text mining and therefore they have considerable noise. Our optimization of the class embeddings proves to be an effective way to reduce this noise and learn robust semantic representations. 

ZSD has numerous applications in novel object localization, retrieval, tracking, and reasoning about object's relationships with its environment using only available semantics, e.g., an object name or a natural language description. Although a critical problem, ZSD is remarkably difficult compared to its classification counterpart. While the zero-shot recognition problem assumes only a single primary object in an image and attempts to predict its category, the ZSD task has to predict both the multi-class category label and precise location of each instance in the given image. Since there can be a prohibitively huge number of possible locations for each object in an image and because the semantic class descriptions are noisy, a detection approach is much more susceptible to incorrect predictions compared to classification. Therefore, it would be expected that a ZSD method predicts a class label that might be incorrect but visually and semantically similar to the corresponding true class. For example, wrongly predicting a `spider' as `scorpion' where both are semantically similar because of being invertebrates. To address this issue, we relax the original detection problem to independently study the confusions emanating from the visual and semantic resemblance between closely linked classes. For this purpose, alongside the ZSD, we evaluate on zero-shot meta-class detection, zero-shot tagging, and zero-shot meta class tagging.  Notably, the proposed network is trained only `once' for ZSD task and the additional tasks are used during evaluations only.

Although deep network based solutions have been proposed for zero-shot recognition \cite{DeViSE_NIPS_2013,Ba_CVPR_2015,Zhang_2017_CVPR}, to the best of our knowledge, we propose the first end-to-end trainable network for the ZSD problem that concurrently relates visual image features with the semantic label information. This network considers semantic embedding vector of classes as a fixed embedding within the network to produce prediction scores for both seen and unseen classes. We propose a novel loss formulation that incorporates max-margin learning \cite{Zhang_2016_CVPR} and a semantic clustering loss based on class-scores of different meta-classes. While the max-margin loss tries to separate individual classes, semantic clustering loss tries to reduce the noise in semantic vectors by positioning similar classes together and dissimilar classes far apart. Notably, our proposed formulation assumes predefined unseen classes to explore the semantic relationships during model learning phase. This assumption is consistent with recent efforts in the literature which consider class semantics to solve the domain shift problem in ZSL \cite{Deng_ECCV_2014,Fu_PAMI_2017} and does not a constitute transductive setting \cite{Deutsch_2017_CVPR,Fu_Transductive_2015,Kodirov_2015_ICCV}. Based on the premise that unseen class semantics may be unknown during training in several practical zero-shot scenarios, we also propose a variant of our approach that can be trained without predefined unseen classes. 
Finally, we propose a comparison method for ZSD by extending a popular zero-shot recognition framework named ConSE \cite{norouzi_arXiv_2013} using Faster-RCNN \cite{Faster_RCNN_2017}. 

In summary, this paper reports the following advances:
\vspace{-0.4em}
\begin{itemize}
\item We introduce a new problem for zero-shot learning, which aims to jointly recognize and localize novel objects in complex scenes. 
\item We present a new experimental protocol and design a novel baseline solution extended from conventional recognition to the detection task.
\item We propose an end-to-end trainable deep architecture that simultaneously considers both visual and semantic information.
\item We design a novel loss function that achieves synergistic effects for max-margin class separation and semantic clustering based on meta-classes. Beside that, our approach can automatically tune noisy semantic embeddings.
\end{itemize}

\section{Problem Description} \vspace{-0.5em}
Given a set of images for seen object categories, ZSD aims at the \emph{recognition} and \emph{localization} of previously unseen object categories. In this section, we formally describe the ZSD problem and its associated challenges. We also introduce variants of the detection task, which are natural extensions of the original problem. First, we describe the notations used in the following discussion. 

\textbf{Preliminaries:}
Consider a set of `seen' classes denoted by $\mathcal{S} = \{1,\ldots, \mathrm{S}\}$, whose examples are available during the training stage and $\mathrm{S}$   represents their total number. There exists another set of `unseen' classes $\mathcal{U} = \{\mathrm{S}+1, \ldots,\mathrm{S}+\mathrm{U}\}$, whose instances are only available during the test phase. We denote the set of all object classes by $\mathcal{C} = \mathcal{S} \cup \mathcal{U} $, such that $\mathrm{C} = \mathrm{S} + \mathrm{U}$ denote the cardinality of the label space. 

We define a set of meta (or super) classes by grouping similar object classes into a single meta category. These meta-classes are  denoted by $\mathcal{M} = \{z_m \, :\, m \in [1,\mathrm{M}]\}$, where $\mathrm{M}$ denote the total number of meta-classes and $z_m = \{k\in \mathcal{C} \; s.t., \, g(k) = m\}$.  Here, $g(k)$ is a mapping function which maps each class $k$ to its corresponding meta-class $z_{g(k)}$. Note that the meta-classes are mutually exclusive i.e., $\cap_{m=1}^{\mathrm{M}} z_m = \phi$ and $\cup_{m=1}^{\mathrm{M}} z_m = \mathcal{C}$.

The set of all training images is denoted by $\mathcal{X}^s$, which contains examples of all seen object classes. The set of all test images containing samples of unseen object classes is denoted by $\mathcal{X}^u$. Each test image $\mathbf{x} \in \mathcal{X}^u$ contains at least one instance of an unseen class. Notably, no unseen class object is present in $\mathcal{X}^s$, but $\mathcal{X}^u$ may contain seen objects. 

We define a $d$ dimensional word vector $\mathbf{v_c}$ (word2vec or GloVe)  for every class $c \in \mathcal{C}$. The ground-truth label for an $i^{th}$ bounding box is denoted by $y_i$. The object detection task also involves identifying the background class for negative object proposals, we introduce the extended label sets: $\mathcal{S}' = \mathcal{S} \cup y_{bg}$, $\mathcal{C}' = \mathcal{C} \cup y_{bg}$ and $\mathcal{M}' = \mathcal{M} \cup y_{bg}$, where $y_{bg} =\{\mathrm{C}+1\}$ is a singleton set denoting the background label.

\textbf{Task Definitions:}
Given the observed space of images $\mathcal{X} = \mathcal{X}^s \cup \mathcal{X}^u$ and the output label space $\mathcal{C}'$, our goal is to learn a mapping function $f : \mathcal{X} \mapsto \mathcal{C}'$ which gives the minimum regularized empirical risk ($\mathcal{\hat{R}}$) as follows:
\begin{align}
\arg \underset{f \in \mathcal{F}}{\min} \;\mathcal{\hat{R}}(f(\mathbf{x};\Theta)) + \Omega(\Theta),
\end{align}
where, $\mathbf{x} \in \mathcal{X}^s$ during training, $\Theta$ denotes the set of parameters and $\Omega(\Theta)$ denotes the regularization on the learned weights. The mapping function has the following form:
\begin{align}\label{eq:comp_fun}
f(\mathbf{x};\Theta) = \arg \underset{y \in \mathcal{C}}{\max} \;\underset{b \in \mathcal{B}(\mathbf{x})}{\max} \mathcal{F}(\mathbf{x}, y, b;\Theta),
\end{align}
where $\mathcal{F}(\cdot)$ is a compatibility function, $\mathcal{B}(\mathbf{x})$ is the set of all bounding box proposals in a given image $\mathbf{x}$. Intuitively, Eq.~\ref{eq:comp_fun} finds the best scoring bounding boxes for each object category and assigns them the maximum scoring object category.
Next, we define the zero-shot learning tasks which go beyond a single unseen category recognition in images. Notably, the training is framed as the challenging ZSD problem, however the remaining task descriptions are used during evaluation to relax the original problem: 
\begin{itemize} 
\item[\bf T1] \textit{Zero-shot detection (ZSD):} Given a test image $\mathbf{x} \in \mathcal{X}^u$, the goal is to categorize and localize each instance of an unseen object class $u \in \mathcal{U}$. 
\item[\bf T2] \textit{Zero-shot meta-class detection (ZSMD):} Given a test image $\mathbf{x} \in \mathcal{X}^u$, the goal is to localize each instance of an unseen object class $u \in \mathcal{U}$ and categorize it into one of the super-classes $m \in \mathcal{M}$.
\item[\bf T3] \textit{Zero-shot tagging (ZST):} To recognize one or more unseen classes in a test image $\mathbf{x} \in \mathcal{X}^u$,  without identifying their location.
\item[\bf T4]  \textit{Zero-shot meta-class tagging (ZSMT):} To recognize one or more meta-classes in a test image $\mathbf{x} \in \mathcal{X}^u$, without identifying their location.
\end{itemize}

Among the above mentioned tasks, the ZSD is the most difficult problem and difficulty level decreases as we go down the list. The goal of the later tasks is to distill the main challenges in ZSD by investigating two ways to relax the original problem:
\textbf{(a)} The effect of reducing the unseen object classes by clustering similar unseen classes into a single super-class (T2 and T4). 
\textbf{(b)} The effect of removing the localization constraint. To this end we investigate the zero-shot tagging problem, where the goal is to only recognize all object categories in an image (T3 and T4). 

The state-of-the-art in zero-shot learning deals with only recognition/tagging. The proposed problem settings add the missing detection task which indirectly encapsulates traditional recognition and tagging task.

\begin{figure}[t] 
  \centering
   \includegraphics[width=.9\textwidth,trim={.2cm 0cm 2.1cm 0cm},clip]{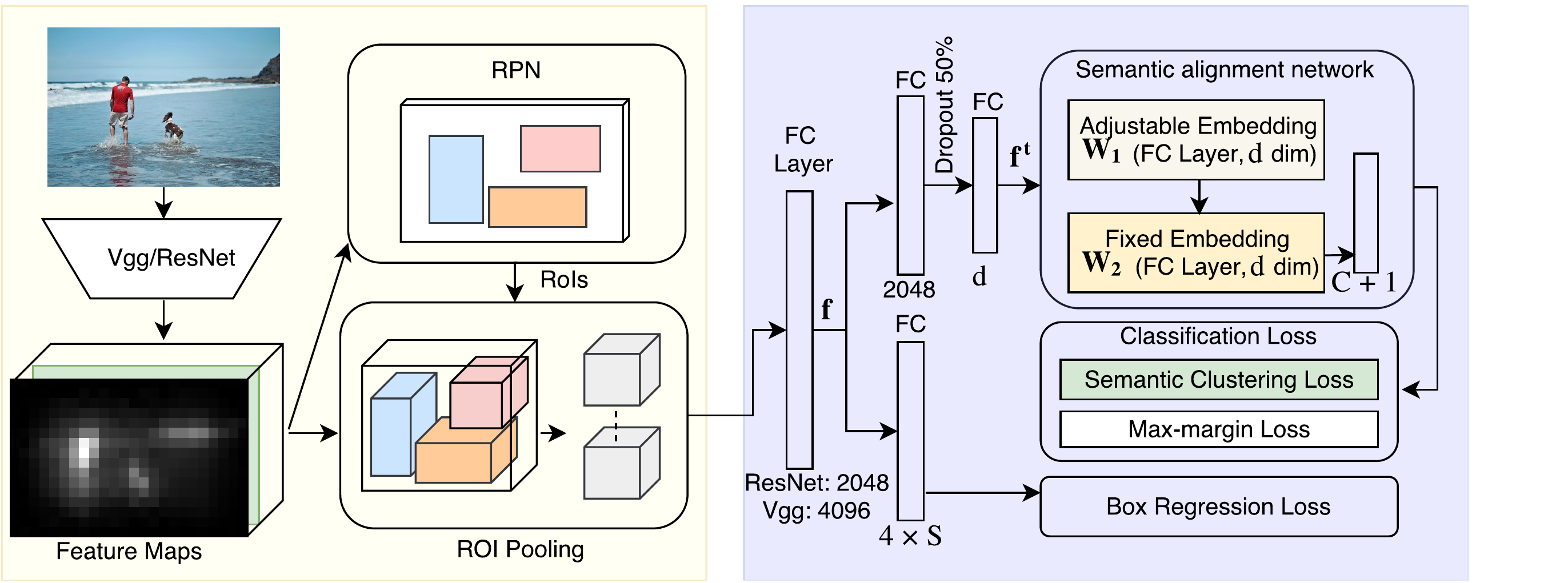}
   \vspace{-1.0em}
   \caption{Network Architecture - \emph{Left:} Image level feature maps are used to propose candidate object boxes and their corresponding features. \emph{Right:} The features are used for classification and localization of new classes by utilizing their semantic concepts.}
\label{fig:network}
\end{figure}

\section{Zero-Shot Detection} \vspace{-0.5em}
   Our proposed model uses Faster-RCNN \cite{Faster_RCNN_2017} as a backbone architecture, due to its superior performance among competitive end-to-end detection models \cite{Dai_RFCN_2016,Liu_SSD_2016,Redmon_yolo9000_2016}. We first provide an overview of our proposed model architecture and then discuss network learning. Finally, we extend a popular ZSL approach to the detection problem, against which we compare our performance in the experiments.
    
\subsection{Model Architecture}
The overall architecture is illustrated in Fig~\ref{fig:network}. It has two main components marked in color: the first provides object-level feature descriptions and the second integrates visual information with the semantic embeddings to perform zero-shot detection. We explain these in detail next.  

\paragraph{Object-level Feature Encoding:}
For an input image $\mathbf{x}$, a deep network (VGG or ResNet) is used to obtain the intermediate convolutional activations. These activations are treated as feature maps, which are forwarded to the Region Proposal Network (RPN). The RPN generates a set of candidate object proposals by automatically ranking the anchor boxes at each sliding window location. The high-scoring candidate proposals can be of different sizes, which are mapped to fixed sized representation using a RoI pooling layer which operates on the initial feature maps and the proposals generated by the RPN. The resulting object level features for each candidate are denoted as `$\mathbf{f}$'. Note that the RPN generates object proposal based on the objectness measure. Thus, a trained RPN on seen objects can generate proposals for unseen objects also. In the second block of our architecture, these feature representations are used alongside the semantic embeddings to learn useful representations for both the seen and unseen object-categories.

\paragraph{Integrating Visual and Semantic Contexts:}
The object-level feature $\mathbf{f}$ is forwarded to two branches in the second module. The \textbf{top branch} is trained to predict the object category for each candidate box. Note that this can assign a class $c\in \mathcal{C}'$, which can be a seen, unseen or background category. The branch consists of two main sub-networks, which are key to learning the semantic relationships between seen and unseen object classes. 

The first component is the `\emph{Semantic Alignment Network}' (SAN), which consist of an adjustable FC layer, whose parameters are denoted as $\mathbf{W}_1 \in \mathbb{R}^{d\times d}$, that projects the input visual feature vectors to a semantic space with $d$ dimensions. The resulting feature maps are then projected onto the \textbf{fixed} semantic embeddings, denoted by $\mathbf{W}_2 \in \mathbb{R}^{d \times (\mathrm{C}+1)}$, which are obtained in an unsupervised manner by text mining (e.g., Word2vec and GloVe embeddings). Note that, here we consider both seen and unseen semantic vectors which require unseen classes to be predefined. This consideration is inline with a very recent effort \cite{Fu_PAMI_2017} which adopt this setting to explore the cluster manifold structure of the semantic embedding space and address domain shift issue.
Given a feature representation input to SAN in the top branch, $\mathbf{f}^t$, the overall operation can be represented as:
    \begin{equation}
		\mathbf{o} = (\mathbf{W_1}\mathbf{W_2})^T \mathbf{f}^t.
    \end{equation}
Here, $\mathbf{o}$ is the output prediction score. The $\mathbf{W_2}$ is formed by stacking semantic vectors for all classes, including the background class. For background class, we use the mean word vectors $\mathbf{v}_b=\frac{1}{\mathrm{C}}\sum_{c=1}^{\mathrm{C}} \mathbf{v}_c$ as its embedding in $\mathbf{W_2}$. 

Notably, a non-linear activation function is not applied between the adjustable and fixed semantic embeddings in the SAN. Therefore, the two projections can be understood as a single learnable projection on to the semantic embeddings of object classes. This helps in automatically updating the semantic embeddings to make them compatible with the visual feature domain. It is highly valuable because the original semantic embeddings are often noisy due to the ambiguous nature of closely related semantic concepts and the unsupervised procedure used for their calculation. In Fig.~\ref{fig:w2vtsne}, we visualize modified embedding space when different loss functions are applied during training.

The \textbf{bottom branch} is for bounding box regression to add suitable offsets to the proposals to align them with the ground-truths such that the precise location of objects can be predicted. This branch is set up in the same manner as in Faster-RCNN \cite{Faster_RCNN_2017}.

\subsection{Training and Inference}\label{sec:training}
We follow a two step training approach to learn the model parameters. The \textbf{first} part involves training the backbone Faster-RCNN  for only seen classes using the training set $\mathcal{X}^s$. This training involves initializing weights of shared layers with a pre-trained Vgg/ResNet model, followed by learning the RPN, classification and detection networks. In the \textbf{second} step, we modify the Faster-RCNN model by replacing the last layer of Faster-RCNN classification branch with the proposed semantic alignment network and an updated loss function (see Fig.~\ref{fig:network}). While rest of the network weights are used from the first step, the weights $\mathbf{W_1}$ are randomly initialized and the $\mathbf{W_2}$ are fixed to semantic vectors of the object classes and not updated during training. 

While training in second step, we keep the shared layers trainable but fix the layers specific to RPN since the object proposals requirements are not changed from the previous step.  The same seen class images $\mathcal{X}^s$  are used for training, consistent with the first step. 
For each given image, we obtain the output of RPN which consists of a total of `$\mathrm{R}$' ROIs belonging to both positive and negative object proposals. 
Each proposal has a corresponding ground-truth label given by $y_i \in \mathcal{S}'$. Positive proposals belong to any of the seen class $\mathcal{S}$ and negative proposals contain only background. In our implementation, we use an equal number of positive and negative proposals. Now, when object proposals are passed through ROI-Pooling and subsequent dense layers, a feature representation $\mathbf{f}_i$ is calculated for each ROI. This feature is forwarded to two branches, the classification branch and regression branch. The overall loss is the summation of the respective losses in these two branches, i.e., classification loss and bounding box regression loss.
\begin{equation*}
L(\mathbf{o}_i,b_i, y_i, b_i^{*}) = 
\underset{\Theta}{\arg\min} \frac{1}{\mathrm{T}} \sum_i \Big( L_{cls}(\mathbf{o}_i,y_i) +   L_{reg}(b_i,b_i^{*}) \Big)
\end{equation*} 
where $\Theta$ denotes the parameters of the network, $\mathbf{o}_i$ is the classification branch output, $\mathrm{T} = \mathrm{N}\times \mathrm{R}$ represents the total number of ROIs in the training set with $\mathrm{N}$ images. $b_i$ and $b_i^{*}$ are parameterized coordinates of predicted and ground-truth bounding boxes respectively and $y_i$ represents the true class label of the $i^{th}$ object proposal.

\begin{SCfigure}[][t]
   \includegraphics[width=.6\linewidth]{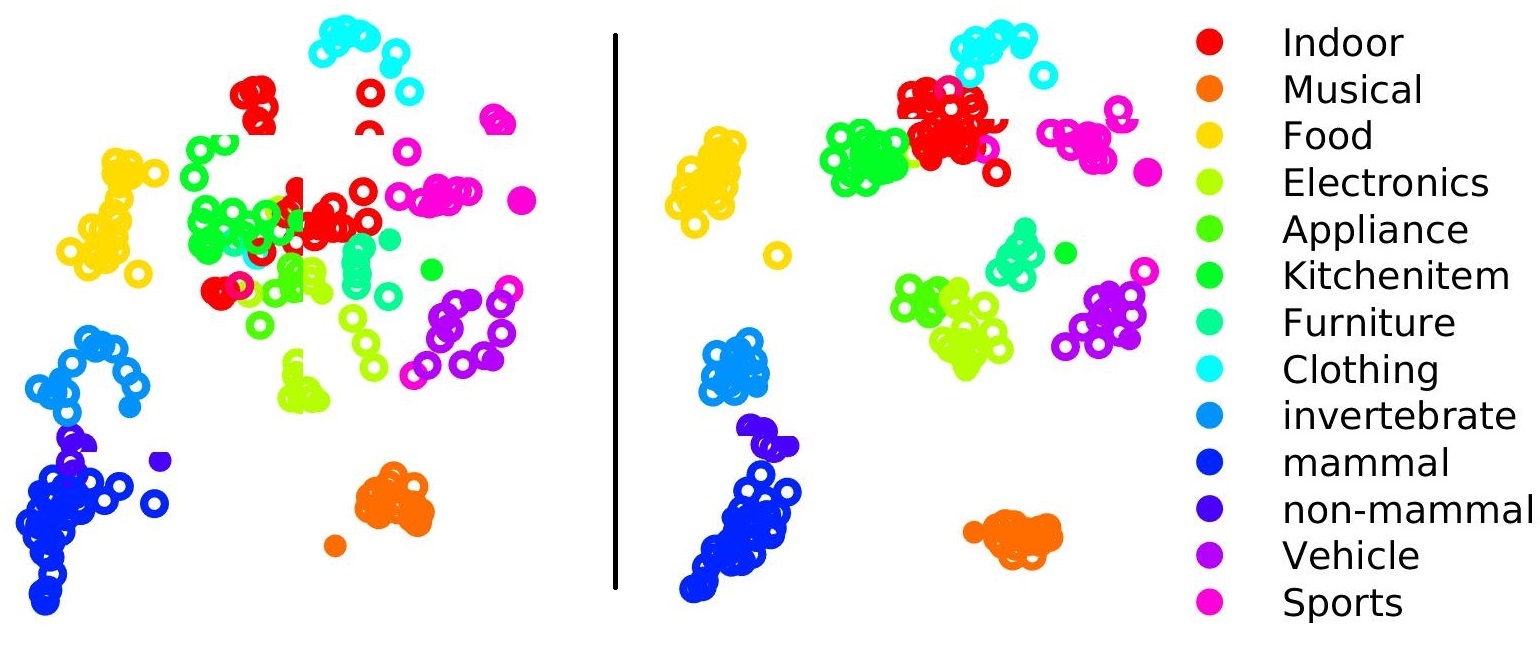}
  \vspace{-0em}
   \caption{The 2D tSNE embedding of modified word vectors $\mathbf{W_1} \mathbf{W_2}$ using only max-margin loss, $L_{mm}$ (left) and with clustering loss, $L_{mm} + L_{mc}$ (right). Semantically similar classes are embedded more closely in cluster based loss.}
\label{fig:w2vtsne}
\end{SCfigure}

\textbf{Classification loss:} This loss deals with both seen and unseen classes. It has two components: a max-margin loss ($L_{mm}$) and a meta-class clustering loss ($L_{mc}$).
\begin{align}
L_{cls}(\mathbf{o}_i,y_i) = \lambda L_{mm}(\mathbf{o}_i,y_i) + (1-\lambda) L_{mc}(\mathbf{o}_i,g(y_i)),
\end{align}
where, $\lambda$ is a hyper-parameter that controls the trade-off between the two losses. We define,
\begin{align} 
L_{mm}(\mathbf{o}_i, y_i) &= \frac{1}{|\mathcal{C}'\setminus y_i|}\sum_{c \in \mathcal{C}'\setminus y_i} \log\Big(1 + \exp  (o_c - o_{y_i})  \Big), \text{ and } \notag\\
L_{mc}(\mathbf{o}_i,g(y_i)) &= \frac{1}{|\mathcal{M}' \setminus z_{g(y_i)}| |z_{g(y_i)}|}\sum_{c \in \mathcal{M}' \setminus z_{g(y_i)}} \sum_{j \in z_{g(y_i)}} \log\Big(1 + \exp   (o_c - o_j) \Big) \notag
\end{align}
where, $o_k$ represents the prediction response of class $k \in \mathcal{S}$. 
$L_{mm}$ tries to separate the prediction response of true class from rest of the classes. In contrast, $L_{mc}$ tries to cluster together the members of each super-class and pulls further apart the classes belonging to different meta-classes. 

We illustrate the effect of clustering loss on the learned embeddings in Fig.~\ref{fig:w2vtsne}. The use of $L_{mc}$ enables us to cluster semantically similar classes together which results in improved embeddings in the semantic space. For example, all animals related meta-classes are in close position whereas food and vehicle are far apart. Such a clear separation in semantic space helps in obtaining a better ZSD performance. Moreover, meta-class based clustering loss does not harm fine-grained detection because the hype-parameter $\lambda$ is used to put more emphasis on the max-margin loss ($L_{mm}$) as compared to the clustering part  ($L_{mc}$) of the overall loss ($L_{cls}$). Still, the clustering loss provides enough guidance to the noisy semantic embeddings (e.g., unsupervised w2v/glove) such that similar classes are clustered together as illustrated in Fig.~\ref{fig:w2vtsne}. Note that w2v/glove try to place similar words nearby with respect to millions of text corpus, it is therefore not fine-tuned for just 200 class recognition setting.


\textbf{Regression loss:} This part of the loss is similar to faster-RCNN regression loss which fine-tunes the bounding box for each seen class ROI. For each $\mathbf{f}_i$, we get $4\times\mathrm{S}$ values representing 4 parameterized co-ordinates of the bounding box of each object instance. The regression loss is calculated based on these co-ordinates and parameterized ground truth co-ordinates. During training, no bounding box prediction is done for background and unseen classes due to unavailability of visual examples. As an alternate approach, we approximate the bounding box for an unseen object through the box proposal for a closely related seen object that achieves maximum response. This is a reasonable approximation because visual features of unseen classes are related to that of similar seen classes. 

\textbf{Prediction:} We normalize each output prediction value of classification branch using
$\hat{o_c} = \frac{o_c}{\parallel \mathbf{v}_c\parallel_2 \parallel \mathbf{f}^{t} \parallel_2}$. It basically calculates the cosine similarity between modified word vectors and image features. This normalization maps the prediction values within 0 to 1 range. We classify an object proposal  as background if maximum responds among $\hat{o_c}$ where $c \in \mathcal{C}'$ belongs to $y_{bg}$. Otherwise, we detect an object proposal as unseen object if its maximum prediction response among $\hat{o_u}$ where $u \in \mathcal{U}$ is above a threshold $\alpha$.
\begin{equation}
		y_u = \arg \max_{u \in \mathcal{U}} \hat{o_u} \quad s.t.,\; \hat{o_u} > \alpha.
\end{equation}
The other detection branch finds $b_i$ which is the parameterized co-ordinates of bounding boxes corresponds to $\mathrm{S}$ seen classes. Among them, we choose a bounding box corresponding to the class having the maximum prediction response in $\hat{o_s}$ where $s \in \mathcal{S}$ for the classified unseen class $y_u$. For the tagging tasks, we simply use the mapping function $g(.)$ to assign a meta-class for any unseen label.

\subsection{ZSD without Pre-defined Unseen} \label{sec:predefined}
While applying clustering loss in Sec.~\ref{sec:training}, the meta-class assignment adds high-level supervision in the semantic space. While doing this assignment, we consider both seen and unseen classes. Similarly, the max-margin loss considers the set $\mathcal{C}'$ consisting of both seen and unseen classes. This problem setting helps to identify the clustering structure of the semantic embeddings to address domain adaptation for zero-shot detection. However, in several practical scenarios, unseen classes may not be known during training. Here, we report a simplified variant of our approach to train the proposed network without pre-defined unseen classes.

For this problem setting, we use only seen+bg word vectors (instead of seen+unseen+bg vectors) as the fixed embedding $\mathbf{W}_2 \in \mathbb{R}^{d \times (\mathrm{S}+1)}$ to train the whole framework with only the max-margin loss, $L'_{mm}$, defined as follows:
$ L'_{mm}(\mathbf{o}_i, y_i) = \frac{1}{|\mathcal{S}'\setminus y_i|}\sum_{c \in \mathcal{S}'\setminus y_i} \log\Big(1 + \exp  (o_c - o_{y_i})  \Big)$. Since the output classification layer cannot make predictions for unseen classes, we apply a procedure similar to ConSE during the testing phase \cite{norouzi_arXiv_2013}. The choice of \cite{norouzi_arXiv_2013} here is made due to two main reasons: \textbf{(a)} In contrast to other ZSL methods which train separate models for each class \cite{Changpinyo_2016_CVPR,rahman2017unified}, ConSE can work on the prediction score of a single end-to-end framework. \textbf{(b)} It is straight-forward to extend a single network to ZSD along with ConSE, since \cite{norouzi_arXiv_2013} uses semantic embeddings only during the test phase. 

Suppose, for an object proposal, $\mathbf{o} \in \mathbb{R}^{\mathrm{S}+1}$ is the vector containing final probability values of only seen classes and background. As described earlier, we ignore the object proposal if the background class get highest probability score. For other cases, we sort the vector $\mathbf{o}$ in descending order to compute a list of indices ${\mathbf{l}}$ and the sorted list $\hat{\mathbf{o}}$:
\begin{align}
\hat{\mathbf{o}} , {\mathbf{l}} = \text{sort(}\mathbf{o}\text{)} \quad s.t., \; {o}_j = \hat{o}_{l_j}.
\end{align}
Then, top $K$ score values (s.t., $K \leq \mathrm{S}$) from $\hat{\mathbf{o}}$ are combined with their corresponding word vectors using the equation:
$
\mathbf{e}_i = \sum_{k=1}^K \hat{\mathbf{o}}_k \mathbf{v}_{l_k}.
$
We consider $\mathbf{e}_i$ as a semantic space projection of an object proposal which is a combination of word vectors weighted by top $K$ seen class probabilities. The final prediction is made by finding the maximum cosine similarity among $\mathbf{e}_i$ and all unseen word vectors,
$$
y_u = \arg \max_{u \in \mathcal{U}} \cos(\mathbf{e}_i,\mathbf{v}_u).
$$
In this paper, we use $K=10$ as proposed in \cite{norouzi_arXiv_2013}. For bounding box detection, we choose the box for which corresponding seen class gets maximum score.

\section{Experiments} \vspace{-0.5em}
	\subsection{Dataset and Experiment Protocol} \label{exp_protocol}    
    
\textbf{Dataset:} We evaluate our approach on the standard ILSVRC-2017 detection dataset \cite{ILSVRC_2015}. This dataset contains 200 object categories. For training, it includes 456,567 images and 478,807 bounding box annotations around object instances. The validation dataset contains 20,121 images fully annotated with the 200 object categories which include 55,502 object instances. A category hierarchy has been defined in \cite{ILSVRC_2015}, where some objects have multiple parents. Since, we also evaluate our approach on meta-class detection and tagging, we define a single parent for each category (see supplementary material for detail).

\textbf{Seen/unseen split:} 
Due to lack of an existing ZSD protocol, we propose a challenging seen/unseen split for ILSVRC-2017 detection dataset. Among 200 object categories, we randomly select 23  categories as unseen and rest of the 177 categories are considered as seen. This split is designed to follows the following practical considerations: \emph{(a)} unseen classes are rare, \emph{(b)} test categories should be diverse, \emph{(c)} the unseen classes should be semantically similar with at least some of the seen classes. The details of split are provided in supplementary material. 

\textbf{Train/test set:} A zero-shot setting does not allow any visual example of an unseen class during training. Therefore, we customize the training set of ILSVRC such that images containing any unseen instance are removed. This results in a total of 315,731 training images with 449,469 annotated bounding boxes. For testing, the traditional zero-shot recognition setting is used which considers only unseen classes. As the test set annotations are not available to us, we cannot separate unseen classes for evaluation. Therefore, our test set is composed of the left out data from ILSVRC training dataset plus validation images having at least one unseen bounding box. The resulting test set has 19,008 images and 19,931 bounding boxes.

\textbf{Semantic embedding:} Traditionally ZSL methods report performance on both supervised attributes and unsupervised word2vec/glove as semantic embeddings. As manually labeled supervised attributes are hard to obtain, only small-scale datasets with these annotations are available \cite{aPY_2009,AwA_2009}. ILSVRC-2017 detection dataset used in the current work is quite huge and does not provide attribute annotations. In this paper, we work on $\ell_2$ normalized 500 and 300 dimensional unsupervised word2vec \cite{Mikolov_NIPS_2013} and GloVe \cite{Jeffrey_Glove_2014} vector respectively to describe the classes. These word vectors are obtained by training on several billion words from Wikipedia dump corpus.

\textbf{Evaluation Metric:} We report average precision (AP) of individual unseen classes and mean average precision (mAP) for the overall performance of unseen classes. 

\textbf{Implementation Details}: Unlike Faster-RCNN, our first step is trained in one step: after initializing shared layer with pre-trained weights, RPN and detection network of Fast-RCNN layers are learned together. Some other settings includes rescaling shorter size of image as 600 pixels, RPN stride = 16, three anchor box scale 128, 256 and 512 pixels, three aspect ratios 1:1, 1:2 and 2:1, non-maximum suppression (NMS) on proposals class probability with IoU threshold = 0.7. Each mini-batch is obtained from a single image having 16 positive and 16 negative (background) proposals. Adam optimizer with learning rate $10^{-5}$, $\beta_1 = 0.9$ and $\beta_2 = 0.999$ is used in both state training. First step is trained over 10 million mini-batches without any data augmentation, but data augmentation through repetition of object proposals is used in second step (details in supplementary material). During testing, the prediction score threshold was 0.1 for baseline and Ours (with $L'_{mm}$) and 0.2 for clustering method (Ours with $L_{cls}$). We implement our model in \textit{Keras}.

\begin{table}[!t]
  \begin{center}
  \scalebox{0.8}{
    \begin{tabular}{|c|c|c|c|c|c|c|c|c|c|c|c|c|}
    \hline
    \multirow{2}{*}{Network}&\multicolumn{3}{c}{ZSD}&\multicolumn{3}{|c|}{ZSMD}&\multicolumn{3}{c}{ZST}&\multicolumn{3}{|c|}{ZSMT} \\ \cline{2-13}
    &\rotatebox{0}{Baseline}&\rotatebox{0}{\parbox{0.9cm}{ Ours ($L'_{mm}$)}}&\rotatebox{0}{\parbox{0.9cm}{Ours ($L_{cls}$)}}&\rotatebox{0}{Baseline}&\rotatebox{0}{\parbox{0.9cm}{ Ours ($L'_{mm}$)}}&\rotatebox{0}{\parbox{0.9cm}{Ours ($L_{cls}$)}}&\rotatebox{0}{Baseline}&\rotatebox{0}{\parbox{0.9cm}{\raggedright Ours ($L'_{mm}$)}}&\rotatebox{0}{\parbox{0.9cm}{\raggedright Ours ($L_{cls}$)}}&\rotatebox{0}{Baseline}&\rotatebox{0}{\parbox{0.9cm}{Ours ($L'_{mm}$)}}&\rotatebox{0}{\parbox{0.9cm}{Ours ($L_{cls}$)}}\\ \hline
 
 R+w2v    &12.7&15.0&\textbf{16.0}&13.7&15.4&\textbf{15.4}&23.3&27.5&\textbf{30.0}&28.8&33.4&\textbf{39.3} \\ \hline
 R+glo    &12.0&12.3&\textbf{14.6}&12.9&14.1&\textbf{16.1}&22.3&24.5&\textbf{26.2}&29.2&31.5&\textbf{36.3} \\ \hline
 V+w2v    &10.2&\textbf{12.7}&11.8&11.4&\textbf{12.5}&11.8&23.3&25.6&\textbf{26.2}&29.0&31.3&\textbf{36.0} \\ \hline
 V+glo    & 9.0&10.8&\textbf{11.6}& 9.7&11.3&\textbf{11.8}&20.3&22.9&\textbf{23.9}&27.3&29.2&\textbf{34.2} \\ \hline
    \end{tabular}}
  \end{center}
  \vspace{-0.8em}
  \caption{mAP of the unseen classes. Ours (with $L'_{mm}$) and Ours (with $L_{cls}$) denote the performance without  predefined unseen and with cluster loss respectively (Sec.~\ref{sec:predefined} and Sec.~\ref{sec:training}) . For cluster case, $\lambda=0.8$. }
  \label{tab:overallresult}
  \vspace{-1.5em}
\end{table}

\begin{table}[!t]
  \begin{center}
  \scalebox{0.58}{
    \begin{tabular}{|c|c||c|c|c|c|c|c|c|c|c|c|c||c|c|c|c|c|c|c|c|c|c|c|c|}
    \hline
    
    
\rotatebox{90}{ }&\rotatebox{90}{\makecell{ \textbf{OVERALL}}}&\rotatebox{90}{p.box}&\rotatebox{90}{syringe}&\rotatebox{90}{harmonica}&\rotatebox{90}{maraca}&\rotatebox{90}{burrito}&\rotatebox{90}{pineapple}&\rotatebox{90}{bowtie}&\rotatebox{90}{s.trunk}&\rotatebox{90}{d.washer}&\rotatebox{90}{canopener}&\rotatebox{90}{p.rack}&\rotatebox{90}{bench}&\rotatebox{90}{e.fan}&\rotatebox{90}{iPod}&\rotatebox{90}{scorpion}&\rotatebox{90}{snail}&\rotatebox{90}{hamster}&\rotatebox{90}{tiger}&\rotatebox{90}{ray}&\rotatebox{90}{train}&\rotatebox{90}{unicycle}&\rotatebox{90}{golfball}&\rotatebox{90}{h.bar}\\  \hline   
 
& & \multicolumn{11}{c||}{Similar classes NOT present} & \multicolumn{12}{c|}{Similar classes present}\\  

& & \multicolumn{11}{c||}{ZSD Baseline = 6.3, Ours ($L'_{mm}$) = \textbf{6.5}, Ours ($L_{cls}$) = 4.4} & \multicolumn{12}{c|}{ZSD Baseline = 18.6, Ours ($L'_{mm}$) = 22.7, Ours ($L_{cls}$) = \textbf{27.4}}\\
\hline



\multicolumn{25}{|c|}{Zero-Shot Detection (ZSD)} \\ \hline    
    
Baseline&12.7&0.0&3.9&\textbf{0.5}&0.0&36.3&\textbf{2.7}&1.8 &1.7&\textbf{12.2}&2.7&\textbf{7.0}&1.0&0.6&22.0&19.0&1.9&40.9&\textbf{75.3}&0.3 &28.4&\textbf{17.9}&12.0&4.0 \\
Ours ($L'_{mm}$) &15.0&0.0&\textbf{8.0}&0.2&0.2&\textbf{39.2}&2.3&\textbf{1.9}&\textbf{3.2}&11.7&\textbf{4.8}&0.0&0.0&\textbf{7.1}&23.3&25.7&\textbf{5.0}&\textbf{50.5}&\textbf{75.3}&0.0&44.8&7.8&\textbf{28.9}&\textbf{4.5} \\
Ours ($L_{cls}$)&\textbf{16.4}&\textbf{5.6}&1.0&0.1&0.0&27.8&1.7&1.5 &1.6&7.2 &2.2&0.0&\textbf{4.1}&5.3&\textbf{26.7}&\textbf{65.6}&4.0&47.3&71.5&\textbf{21.5}&\textbf{51.1}&3.7 &26.2&1.2 \\ \hline 

\multicolumn{25}{|c|}{Zero-Shot Tagging (ZST)} \\ \hline 

Baseline&23.3&2.9 &13.4&9.6 &3.1&61.7&20.7&16.3 &7.5&29.4&8.6 &\textbf{12.2}&8.5 &4.9&46.2 &30.7&11.0&51.8&77.6&9.0 &46.1&\textbf{39.0}&12.7&12.6 \\
Ours ($L'_{mm}$)&27.5&2.9&\textbf{20.8}&10.5&3.3&\textbf{72.5}&\textbf{27.7}&16.7&7.9&22.9&\textbf{14.3}&2.8&6.7&\textbf{14.5}&46.8&42.6&\textbf{16.0}&\textbf{59.1}&\textbf{80.0}&12.9&67.3&34.1&\textbf{34.0}&\textbf{17.1}\\
Ours ($L_{cls}$)&\textbf{30.6}&\textbf{12.6}&10.2&\textbf{11.9}&\textbf{4.9}&48.9&21.8&\textbf{17.9}&\textbf{29.1}&\textbf{32.2}&10.0&4.1 &\textbf{20.7}&10.7&\textbf{52.2}&\textbf{82.6}&12.3&58.5&75.5&\textbf{48.9}&\textbf{72.2}&16.9&33.9&15.5 \\ \hline 

\hline \hline

{Meta-class}&{}& \multicolumn{2}{c|}{{Indoor}} &	\multicolumn{2}{c|}{{Musical}}&	\multicolumn{2}{c|}{{Food}}&	\multicolumn{2}{c|}{{Clothing}}&	{Appli.} &	\multicolumn{2}{c||}{{Kitchen}}& {Furn.}&	\multicolumn{2}{c|}{{Electronic}}&	\multicolumn{2}{c|}{{Invertebra.}}&	\multicolumn{2}{c|}{{Mammal}}&	{Fish} &	\multicolumn{2}{c|}{{Vehicle}}&	\multicolumn{2}{c|}{{Sport}} \\ \hline

\multicolumn{25}{|c|}{Zero-Shot Meta Detection (ZSMD)} \\ \hline 

Baseline& 13.7 & \multicolumn{2}{c|}{3.3}&\multicolumn{2}{c|}{\textbf{0.3}}&\multicolumn{2}{c|}{\textbf{24.0}}&\multicolumn{2}{c|}{\textbf{4.0}}&\textbf{12.2}&\multicolumn{2}{c||}{2.1}&1.0&\multicolumn{2}{c|}{12.1}&\multicolumn{2}{c|}{17.0}&\multicolumn{2}{c|}{70.7}&0.3&\multicolumn{2}{c|}{22.1}&\multicolumn{2}{c|}{8.5}  \\

Ours ($L'_{mm}$)&15.4& \multicolumn{2}{c|}{\textbf{8.1}}&\multicolumn{2}{c|}{0.1}&\multicolumn{2}{c|}{18.4}&\multicolumn{2}{c|}{2.3}&11.7&\multicolumn{2}{c||}{\textbf{3.0}}&0.0&\multicolumn{2}{c|}{14.3}&\multicolumn{2}{c|}{27.8}&\multicolumn{2}{c|}{\textbf{73.6}}&0.0&\multicolumn{2}{c|}{\textbf{32.1}}&\multicolumn{2}{c|}{9.0} \\

Ours ($L_{cls}$)& \textbf{15.6}&\multicolumn{2}{c|}{3.5}&\multicolumn{2}{c|}{0.1}&\multicolumn{2}{c|}{10.0}&\multicolumn{2}{c|}{1.9}&7.2&\multicolumn{2}{c||}{1.2}&\textbf{4.1}&\multicolumn{2}{c|}{\textbf{15.3}}&\multicolumn{2}{c|}{\textbf{31.4}}&\multicolumn{2}{c|}{66.8}&\textbf{21.5}&\multicolumn{2}{c|}{31.2}&\multicolumn{2}{c|}{\textbf{9.3}} \\ \hline

\multicolumn{25}{|c|}{Zero-Shot Meta-class Tagging (ZSMT)} \\ \hline 

Baseline& 28.8&\multicolumn{2}{c|}{15.2}&\multicolumn{2}{c|}{12.0}&\multicolumn{2}{c|}{55.6}&\multicolumn{2}{c|}{25.2}&29.4&\multicolumn{2}{c||}{10.7}&8.5&\multicolumn{2}{c|}{31.5}&\multicolumn{2}{c|}{36.5}&\multicolumn{2}{c|}{75.8}&9.0&\multicolumn{2}{c|}{48.4}&\multicolumn{2}{c|}{17.0} \\

Ours ($L'_{mm}$)&33.4& \multicolumn{2}{c|}{\textbf{24.1}}&\multicolumn{2}{c|}{13.6}&\multicolumn{2}{c|}{\textbf{55.9}}&\multicolumn{2}{c|}{31.3}&22.9&\multicolumn{2}{c||}{\textbf{14.7}}&6.7&\multicolumn{2}{c|}{33.0}&\multicolumn{2}{c|}{49.4}&\multicolumn{2}{c|}{82.6}&12.9&\multicolumn{2}{c|}{64.2}&\multicolumn{2}{c|}{23.2} \\

Ours ($L_{cls}$)& \textbf{39.9}&\multicolumn{2}{c|}{19.2}&\multicolumn{2}{c|}{\textbf{15.5}}&\multicolumn{2}{c|}{45.6}&\multicolumn{2}{c|}{\textbf{38.5}}&\textbf{32.2}&\multicolumn{2}{c||}{12.4}&\textbf{20.7}&\multicolumn{2}{c|}{\textbf{40.3}}&\multicolumn{2}{c|}{\textbf{58.2}}&\multicolumn{2}{c|}{\textbf{84.8}}&\textbf{48.9}&\multicolumn{2}{c|}{\textbf{74.7}}&\multicolumn{2}{c|}{\textbf{27.1}} \\ \hline

    \end{tabular}}
  \end{center}
  \vspace{-0.5em}
  \caption{Average precision of individual unseen classes using ResNet+w2v and loss configurations $L'_{mm}$ and $L_{cls}$ (cluster based loss with $\lambda=0.6$). We have grouped unseen classes into two groups based on whether visually similar classes present in the seen class set or not. Our proposed method achieve significant performance improvement for the group where similar classes are present in the seen set.} 
  \vspace{-1.5em}
  \label{tab:individual_map}
\end{table}

 

\subsection{ZSD Performance}

We compare different versions of our method (with loss configurations $L'_{mm}$ and $L_{cls}$ respectively) to a baseline approach. Note that the baseline is a simple extension of Faster-RCNN \cite{Faster_RCNN_2017} and ConSE \cite{norouzi_arXiv_2013}. We apply the inference strategy mentioned in Sec. \ref{sec:predefined} after first step training as we can still get a vector $\mathbf{o} \in \mathbb{R}^{\mathrm{S}+1}$ on the classification layer of Faster-RCNN network. We use two different architectures i.e., VGG-16 (V) \cite{Vgg_arXiv_2014} and ResNet-50 (R) \cite{ResNet_CVPR_2016} as the backbone of the Faster-RCNN during the first training step. In second step, we experiment with both Word2vec and GloVe as the semantic embedding vectors used to define $\mathbf{W}_2$. Fig. \ref{fig:output} illustrates some qualitative ZSD examples. More performance results of ZSD on other datasets is provided in the supplementary material.

\textbf{Overall results:} Table \ref{tab:overallresult} reports the mAP for all approaches on four tasks: ZSD, ZSMD, ZST, and ZSMT across different combinations of network architectures. We can make following observations: \emph{(1)} Our cluster based method outperforms other competitors on all four tasks because its loss utilizes high-level semantic relationships from meta-class definitions which are not present in other methods.
\emph{(2)} Performances get improved from baseline to Ours (with $L'_{mm}$) across all zero-shot tasks. The reason is baseline method did not consider word vectors during the training. Thus, overall detection could not get enough supervision about the semantic embeddings of classes. In contrast, $L'_{mm}$ loss formulation considers word vectors.
\emph{(3)} Performances get improved from ZST to ZSMT across all methods whereas similar improvement is not common from ZSD to ZSMD. It's not surprising because ZSMD can get some benefit if meta-class of the predicted class is same as the meta-class of true class. If this is violated frequently, we cannot expect significant performance improvement in ZSMD. 
\emph{(4)} In comparison of traditional object detection results, ZSD achieved significantly lower performance. Remarkably, even the state-of-the-art zero-shot classification approaches perform quite low e.g., a recent ZSL method \cite{Zhang_2017_CVPR} reported $11\%$ hit@1 rate on ILSVRC 2010/12. This trend does not undermine to significance of ZSD, rather highlights the underlying challenges.

\textbf{Individual class detection:} Performances of individual unseen classes indicate the challenges for ZSD. In Table \ref{tab:individual_map}, we show performances of individual unseen classes across all tasks with our best (R+w2v) network. We observe that the unseen classes for which visually similar classes are present in their meta-classes achieve better detection performance (ZSD mAP 18.6, 22.7, 27.4) than those which do not have similar classes (ZSD mAP 6.3, 6.5, 4.4) for the all methods (baseline, our's with $L'_{mm}$ and $L_{cls}$). Our proposed cluster method with loss $L_{cls}$ outperforms the other versions significantly for the case when visually similar classes are present. For the all classes, our cluster method is still the best (mAP: cluster 16.4 vs. baseline 12.7). However, our's with $L'_{mm}$ method performs better for when case similar classes are not present (mAP 6.5 vs 4.4).
For the easier tagging tasks (ZST and ZSMT), the cluster method gets superior performance in most of the cases. This indicates that one potential reason for the failure cases of our cluster method for ZSD might be confusions during localization of objects due to ambiguities in visual appearance of unseen classes. 

\begin{wrapfigure}{L}{0.5\textwidth}
  \begin{center}
  \vspace{-3em}
   \includegraphics[width=1\linewidth,trim={.87cm 0cm 1.4cm .5cm},clip]{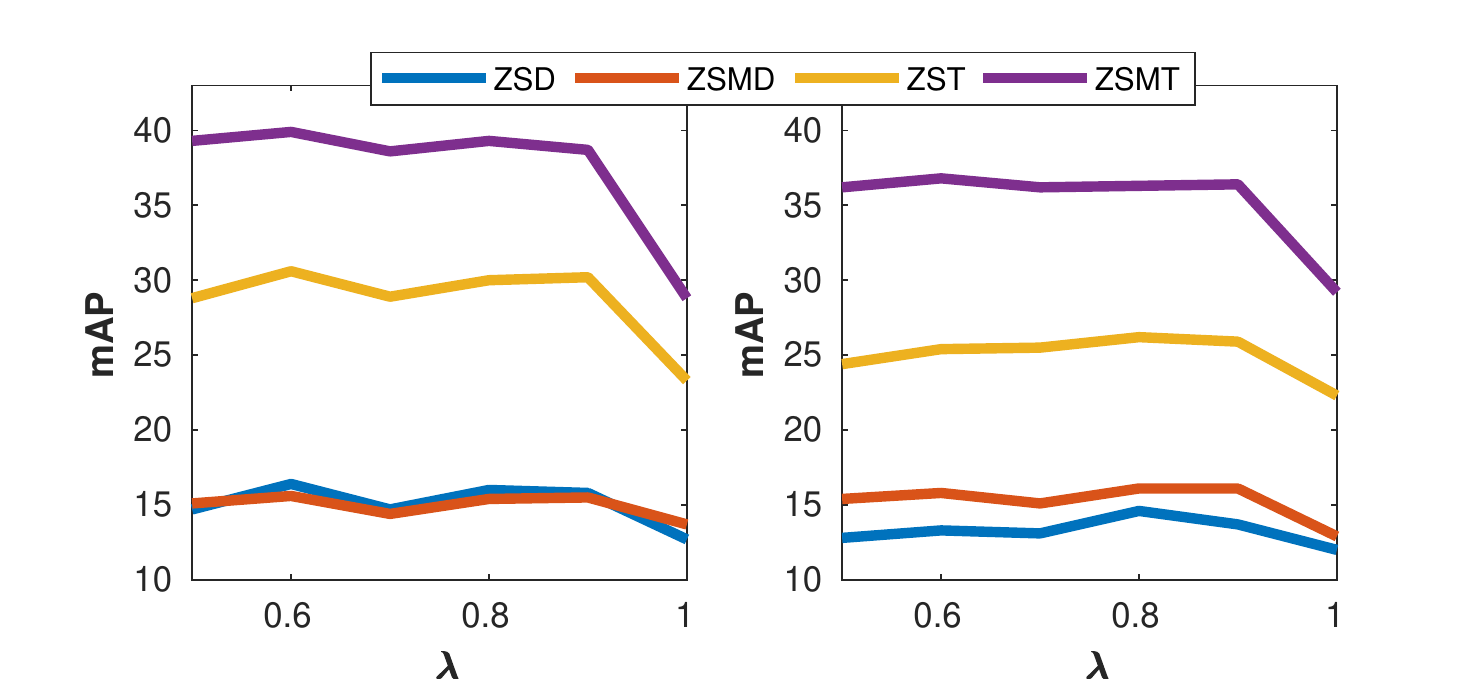}
  \end{center}
  \vspace{-2em}
   \caption{Effect of varying $\lambda$ in different zero-shot tasks for ResNet+w2v (left) and ResNet+glo (right).}
\label{fig:lambdatune}
\vspace{-4.0em}
\end{wrapfigure}

\textbf{Varying $\lambda$:} The hyperparameter $\lambda$ controls the weight between $L_{mm}$ and $L_{mc}$ in $L_{cls}$. In Fig.~\ref{fig:lambdatune}, we illustrate the effect of varying $\lambda$ on four zero-shot tasks for R+w2v and R+glo. It shows that performances has less variation in the range of $\lambda = .5$ to $.9$ than $\lambda = .9$ to $1$. For a larger $\lambda$, mAP starts dropping  since the impact of $L_{mc}$ decreases significantly.

\begin{SCtable}[60][!t]
  \scalebox{.75}{
    \begin{tabular}{|c|c|c|c|}
    \hline
    Top1 Accuracy & Network & w2v  & glo \\
    \hline\hline
    Akata'16 \cite{Akata_2016_CVPR} & V &33.90&- \\
    DMaP-I'17\cite{Li_2017_CVPR} & G+V &26.38&30.34 \\ 
    SCoRe'17\cite{Morgado_2017_CVPR} & G &31.51&- \\
    Akata'15 \cite{Akata_CVPR_2015} & G &28.40&24.20 \\
    LATEM'16 \cite{Xian_2016_CVPR} & G &31.80&32.50 \\
    DMaP-I'17 \cite{Li_2017_CVPR} & G &26.28&23.69 \\
    \hline    
    Ours & R & \textbf{36.77} &\textbf{36.82} \\ 
    \hline
    \end{tabular}}
  \caption{Zero shot recognition on CUB using $\lambda = 1$ because no meta-class assignment is done here. For fairness, we only compared our result with the inductive setting of other methods without per image part annotation and description. We refer V=VGG, R=ResNet, G=GoogLeNet.
}
  \label{tab:cub_shorter}
\end{SCtable}

    \subsection{Zero Shot Recognition (ZSR)}
    Being a detection model, the proposed network can also perform traditional ZSR. We evaluate ZSR performance on popular Caltech-UCSD Birds-200-2011 (CUB) dataset \cite{CUB_2011}. This dataset contains 11,788 images from 200 classes and provides single bounding boxes per image. Following standard train/test split \cite{Xian_CVPR_2017}, we use 150 seen and 50 unseen classes for experiments. For semantics embedding, we use 400-d word2vec (w2v) and GloVe (glo) vector \cite{Xian_2016_CVPR}. Note that, we do not use per image part annotation (like \cite{Akata_2016_CVPR}) and descriptions (like \cite{Zhang_2017_CVPR}) to enrich semantic embedding. For a given test image, our network predicts unseen class bounding boxes. We pick only one label with the highest prediction score per image. In this way, we report the mean Top1 accuracy of all unseen classes in Table \ref{tab:cub_shorter}. One can find our proposed solution achieve significant performance improvement in comparison with state-of-the-art methods.

\begin{figure}[t]
 \centering
 \includegraphics[width=1\textwidth,trim={0cm 0cm 0cm 0cm},clip]{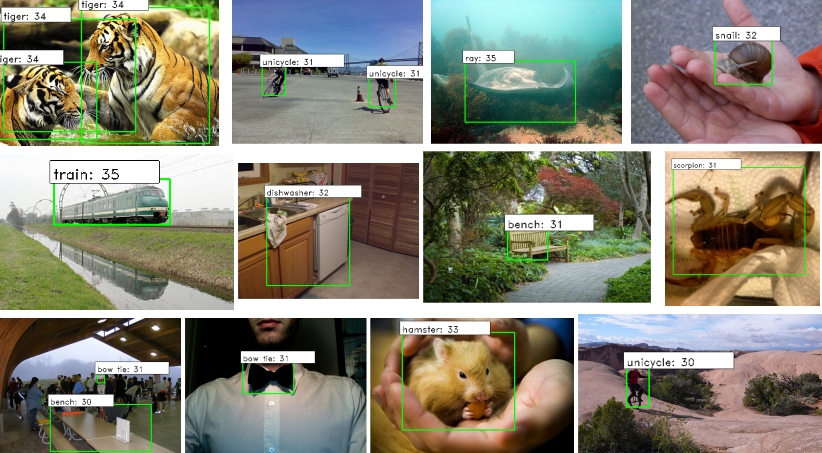}
  \vspace{-2.1em}
   \caption{Selected examples of ZSD of our cluster ($\lambda=.6$)  method with R+w2v, using the prediction score threshold = 0.3. (See supplementary material for more examples) 
   }
\label{fig:output}
\end{figure}

\subsection{Challenges and New Directions}\label{sec:further}
\textbf{ZSD is challenging:} Our empirical evaluations show that ZSD needs to deal with the following challenges: \emph{(1)} Unseen classes are rare compared to seen classes; \emph{(2)} Small unseen objects are hard to detect and harder to relate with their semantics; \emph{(3)} The scarcity of similar seen class leads to an inadequate description of an unseen class; \emph{(4)} As derived in an unsupervised manner, the noise of semantic space affects ZSD. These issues are discussed in detail in supplementary material.

       



\noindent \textbf{Future challenges:} The ZSD problem warrants further investigation. \emph{(1)} Unlink current work one can consider fine-tuning the bounding box of the both seen and unseen classes based on visual and semantic correspondences. \emph{(2)} Rather mapping image feature to the semantic space, the reverse mapping may help ZSD similar to ZSR used in \cite{Kodirov_2017_CVPR,Zhang_2017_CVPR}. \emph{(3)} One can consider the fusion of different word vectors (word2vec and GloVe) to improve ZSD. \emph{(4)} Like generalized ZSL \cite{Xu_Matrix_CVPR_2017,Xian_CVPR_2017,Li_2017_CVPR}, one can extend it to a more realistic generalized ZSD. Moreover, weakly supervised or semi-supervised version of zero shot problems is also possible while performing ZSD/GZSD.

\vspace{-0.5em}
\section{Conclusion}
\vspace{-0.5em}
While traditional ZSL research focuses on only object recognition, we propose to extend the problem to object detection (ZSD). To this end, we offer a new experimental protocol with ILSVRC-2017 dataset specifying the seen-unseen, train-test split. We also develop an end-to-end trainable CNN model to solve this problem. We show that our solution is better than a strong baseline.

Overall, this research throws some new challenges to ZSL community. To make a long-standing progress in ZSL, the community needs to move forward in the detection setting rather than merely recognition. 
 
{\small
\bibliographystyle{ieee}
\bibliography{ref}
}


\title{Supplementary Material}
\author{\large Zero Shot Object Detection}
\institute{}
\maketitle
\def\thesection{\Alph{section}.}
\renewcommand\thesubsection{\Alph{section}.\arabic{subsection}}

\section{Related Work} \label{sec:related_work}
\textbf{End-to-end Object detection:}  Though object detection has been extensively studied in the literature, we can only find a few end-to-end learning pipelines capable of simultaneous object localization and classification. Popular examples of such approaches are Faster R-CNN \cite{Faster_RCNN_2017}, R-FCN \cite{Dai_RFCN_2016}, SSD \cite{Liu_SSD_2016} and YOLO \cite{Redmon_yolo9000_2016}. The contribution of these methods relies on object localization process. Methods like Faster R-CNN \cite{Faster_RCNN_2017}, R-FCN \cite{Dai_RFCN_2016} are based on Region Proposal Network (RPN) which provides bounding box proposals of possible objects and then classifying and fine tuning the box prediction in the later layers. In contrast, methods like SSD \cite{Liu_SSD_2016} and YOLO \cite{Redmon_yolo9000_2016} draw bounding box and classify it in a single step. Unlike RPN; these methods predict bounding box offset of some pre-defined anchors instead of the box co-ordinates itself. The later methods are generally faster than the previous ones. However, RPN based methods are better in terms of accuracy. In current work, we prioritize accuracy over speed. Therefore, we build zero-shot object detection model based on the Faster RCNN.

\textbf{Semantic embedding:} Any zero-shot task like recognition or tagging requires semantic information of classes. This semantic information works as a bridge among seen and unseen classes. The common way to preserve the semantic information of a class is by using a one-dimensional vector. The vector space that holds semantic information of classes is called `semantic embedding space'. Visually similar classes reside in a close position in this space. The semantic vector of any class can be generated both manually or automatically. The manually generated semantic vectors are often called attributes \cite{CUB_2011,Lampert_PAMI_2014}. Although attributes can describe a class with less noise (than other kinds of embeddings), those are very hard to obtain because of manual annotations. As a workaround, automatic semantic embedding can be generated from a large corpus of unannotated text like (Wikipedia, news article, etc.) or hierarchical relationship of classes in WordNet \cite{Wordnet_1995}. Some popular examples of such kind of semantic embeddings are word2vec \cite{Mikolov_NIPS_2013}, GloVe \cite{Jeffrey_Glove_2014}, and hierarchies \cite{Xian_2016_CVPR}. As generated by an unsupervised manner, such embeddings become noisy but provide more flexibility and scalability than manual vectors.

\textbf{Zero-shot learning:} Humans can recognize an object by relating known objects, without prior visual experience. Simulating this behavior into an automated machine vision system is called Zero-shot learning (ZSL). ZSL attempts to recognize unseen objects without any visual examples of the unseen category. In recent years, numerous effective methods for ZSL have been proposed. Every ZSL strategy has to relate seen and unseen embedding through semantic embedding vector. Based on how this relation is established, we can categorize ZSL strategies into three types. The \textbf{first} type of methods attempt to predict the semantic vector of classes \cite{Hinton_NIPS_2009,Wang_CVPR_2013,Lampert_PAMI_2014,Yu_CVPR_2013}. An object is classified as an unseen class based on similarity of predicted vector and semantic vectors of unseen classes. Predicting a high dimensional vector is not an efficient way to related seen-unseen classes because it cannot work consistently if the semantic vectors are noisy \cite{Jayaraman_NIPS_2014}. This reason provokes this kind of methods to use attributes as semantic embedding as they are less noisy. The \textbf{second} kind of methods learn a linear \cite{Akata_PAMI_2016,Akata_CVPR_2015,romera_ICML_2015} or non-linear \cite{Xian_2016_CVPR,Socher_NIPS_2013} compatibility function to relate the seen image feature and corresponding semantic vector. This compatibility function yields high value if visual feature and semantic vector come from the same class and vice versa. A visual feature is classified to an unseen class if it gets the best compatibility score among all possible unseen classes. Such methods work consistently across a wide variety of semantic embedding vectors. The \textbf{third} kind of methods describe unseen classes by mixing seen visual features and semantic embedding \cite{norouzi_arXiv_2013,Changpinyo_2016_CVPR,Zhang_2015_ICCV}. For this mixing purpose, sometimes methods perform per class learning and later combine individual class output to decide outputs for unseen classes. While most of the ZSL approaches convert visual feature to semantic spaces, \cite{Kodirov_2017_CVPR,Zhang_2017_CVPR} mapped semantic vectors to the visual domain to address the hubness problem during prediction \cite{Shigeto_Hubness_2015}. Irrespective of method types, attributes work better as semantic embeddings compared to unsupervised word2vec, GloVe, and hierarchies because of less noise. To minimize this performance gap, researchers have investigated transductive setting \cite{Ye_DSRL2017_CVPR,Xu_Matrix_CVPR_2017,Li_2017_CVPR}, domain adaptation \cite{Deutsch_2017_CVPR,Kodirov_2015_ICCV} and class-attribute association \cite{Al-Halah_2016_CVPR,Demirel_2017_ICCV} techniques. Usually, all ZSL methods are evaluated on a restricted case of recognition problem where test data only contain unseen images. Few recent efforts performed experiments on generalized version of ZSL \cite{Xu_Matrix_CVPR_2017,Xian_CVPR_2017,Li_2017_CVPR}. They found that established ZSL methods perform poorly in such settings. Still, all these methods perform a recognition task in zero-shot settings. In this paper, we extend recognition problem to a more complex detection problem.

\textbf{Zero-shot image tagging:} Instead of assigning one unseen label to an image during recognition task, zero-shot tagging allows to tag multiple unseen tags to an image and/or ranking the array of unseen tags. Very few papers addressed the zero-shot version of this problem \cite{Li_tagging_2015,Fu_Transductive_2015,Zhang_2016_CVPR}. Li et al. \cite{Li_tagging_2015} applied the idea of \cite{norouzi_arXiv_2013} in tagging. They argued that semantic embeddings (like word2vec) of all possible tags may not be available, and therefore, proposed a hierarchical semantic embedding method for those unavailable tags based on its ancestor classes in WordNet hierarchy. \cite{Fu_Transductive_2015} considered the power set of fixed unseen tags as the label set to perform transductive multi-label learning. Recently, \cite{Zhang_2016_CVPR} proposed a fast zero-shot tagging approach that can rank both seen and arbitrary unseen tags during the testing stage. All previous attempts are not end-to-end because they preform training on the top of pre-trained CNN features. In this paper, we propose an end-to-end method for  seen detection with zero-shot object tagging.

\textbf{Object-level attribute reasoning:} 
Object level attribute reasoning has been studied under two themes in the literature. The first theme advocates the use of object-level semantic representations in a traditional ZSL setting. Li et al. \cite{li2014attributes} proposed to use local attributes and employed these shared characteristics to obtain zero-shot classification and segmentations. However, they dealt with fine-grained categorization task, where both seen and unseen objects have similar shapes (and segmentation masks), there is a single dominant category in each image and work with only supervised attributes. Another approach aiming at zero-shot segmentation is to learn a shape space shared with the novel objects. This technique, however, can only segment new object shapes that are very similar to the training set \cite{jetley2016straight}. Along the second theme, some efforts have more recently been reported for object localization and tracking using natural language descriptions \cite{hu2016natural,li2017tracking}. Different to our problem, they assume an accurate semantic description of the object, use supervised examples of objects during training, and therefore do not tackle the zero-shot detection problem. 

\section{Dataset and Experiment Protocol} \label{sec:dataset}

\subsection{Meta-class assignment} \label{sec:metaclass}

The classes of ILSVRC detection dataset maintain a defined hierarchy \cite{ILSVRC_2015}. However, this hierarchy does not follow a tree structure. In this paper, we choose a total of $\mathrm{M} = 14$ meta-classes (including person), in which the 200 object classes are divided. Table \ref{tab:superclasslist} describes meta-class assignment of all 200 classes. This assignment mostly follows the hierarchy of question prescribed in the original paper \cite{ILSVRC_2015}. Few notable exceptions are (1) the classes of first-aid/medical items, cosmetics, carpentry items, school supplies and bag are grouped as indoor accessory, (2) liquid container related classes are merged with kitchen items, (3) flower pot is considered as furniture similar to MicroSoft COCO super-categories \cite{MSCOCO_2014}, (4) All living organisms (other than people) related classes are grouped into three different meta-class categories based on their similarity in word vector embedding space: invertebrate, mammal and non-mammal animal. Although one can argue that all invertebrate are non-mammal, this is just an assignment definition we apply in this paper to obtain a uniform distribution of images across super-classes.

\begin{table*}[!t]
  \begin{center}
    \begin{tabular}{|c|c|c|}
    \hline
   	ID&Meta/Super-class & Categories  \\
    \hline\hline
1&\makecell{Indoor\\Accessory (25)} &	\makecell{axe,	backpack,	band aid,	binder,	chain saw,	cream,	crutch,\\face-powder,	hairspray,	hammer, lipstick, nail,	neck-brace, \\ \textbf{pencilbox},	pencilsharpener,	perfume,	plastic-bag, \\power-drill, purse, rubber-eraser,	ruler,	screwdriver, \\stethoscope, stretcher, \textbf{syringe}} \\ \hline	

2&Musical (17) & \makecell{accordion,	banjo,	cello,	chime,	drum,	flute,	french-horn,\\guitar, \textbf{harmonica},	harp,	\textbf{maraca},	oboe, piano,	saxophone,	\\trombone,	trumpet,	violin} \\ \hline

3&Food (21)&\makecell{apple,	artichoke,	bagel,	banana,	bell-pepper,	\textbf{burrito},	cucumber,	\\fig,	guacamole,	hamburger,	head-cabbage,	hotdog,	lemon,	\\mushroom,	orange,	\textbf{pineapple},	pizza,	pomegranate,\\	popsicle,	pretzel,	strawberry} \\ \hline

4&Electronics (16)&\makecell{computer-keyboard,computer-mouse,	digital-clock,	\textbf{electric-fan},	\\hair-dryer,	\textbf{iPod},	lamp,	laptop, microphone,	printer, vacuum,\\	remote-control,	tape-player, traffic-light,	tv or monitor, washer} \\ \hline

5&Appliance (7)&\makecell{coffee-maker,	\textbf{dishwasher},	microwave,	refrigerator,	stove,	\\toaster,	waffle-iron} \\ \hline

6&\makecell{Kitchen\\item}(17)&\makecell{beaker,	bowl,	\textbf{can-opener},	cocktail-shaker,	corkscrew,	\\cup or mug,	frying-pan,	ladle,	milk-can,	pitcher,	\textbf{plate-rack},\\	 salt or pepper shaker,	soap-dispenser,	spatula,	strainer,	\\water-bottle,	wine-bottle} \\ \hline

7&Furniture (8)&\makecell{baby-bed,	\textbf{bench}, bookshelf,chair,	filing-cabinet,	flower-pot,	\\sofa,	table} \\ \hline																				
8&Clothing (11)& \makecell{bathing-cap,	\textbf{bow-tie},	brassiere,	diaper,	hat with a wide brim,\\	helmet,	maillot,	miniskirt,	sunglasses, \textbf{swimming-trunks}, tie}	\\ \hline
       
9&\makecell{Invertebrate\\animal} (14)&\makecell{ant,	bee,	butterfly,	centipede,	dragonfly,	goldfish,	isopod,\\	jellyfish,	ladybug,	lobster,	\textbf{scorpion}, \textbf{snail},	starfish,	tick }\\ \hline

10&\makecell{mammal\\animal}(28)&	\makecell{antelope,	armadillo,	bear,	camel,	cattle,	dog,	domestic-cat,\\	elephant,	fox,	giant-panda,	\textbf{hamster},hippopotamus,	horse,	\\koala-bear,	lion,	monkey,	otter,	porcupine,	rabbit,	red-panda,	\\seal,	sheep,	skunk,	squirrel,	swine,	\textbf{tiger},	whale,	zebra} \\ \hline

11&\makecell{non-mammal\\animal}(6)&	bird,	frog,	lizard,	\textbf{ray},	snake,	turtle \\ \hline																				
12&Vehicle(12) & \makecell{airplane,	bicycle,	bus,	car,	cart,	golfcart, motorcycle,	\\snowmobile,	snowplow, \textbf{train},	\textbf{unicycle},	watercraft}	\\ \hline

13&Sports (17) & \makecell{balance-beam,	baseball,	basketball,	bow,	croquet-ball,	dumbbell,\\	\textbf{golf-ball},	\textbf{horizontal-bar},	ping-pong-ball,	puck,	punching-bag,\\	racket,	rugby-ball,	ski,	soccer-ball,	tennis-ball,	volleyball} \\ \hline
14&Person (1) & person \\
    \hline
    \end{tabular}
  \end{center}
  \caption{Assigned meta-class to each of the 200 object categories. The unseen classes are presented as bold.}
  \label{tab:superclasslist}
\end{table*} 

\subsection{Train/Test Split}\label{sec:split}
Since the unseen classes are rare in real life settings and therefore their images are hard to collect, we assume that the training set only contains frequent classes. 
For ILSVRC detection dataset, number of instances per class follows a long-tail distribution (Figure \ref{fig:longtail}). For each of our defined meta-class categories, we first plot the instance distribution of the child classes like Figure \ref{fig:superclsbar}. Then, we randomly select one or two classes (depending on the number of child classes) from the rare second half of the distribution. We choose two unseen classes from the meta-classes which have relatively large (9 or more) number of child classes. In contrast, we choose one class as unseen for the meta-classes having less number of child classes. The only exception is that we do not choose `Person' meta-class as unseen because it has no similar child class.

This random selection procedure avoids biasness, ensures diversity (due to selection from all meta-classes) and conforms to the observation that unseen classes are not frequent. 

\subsection{Data Augmentation} \label{sec:augmentation}

\begin{figure}[t]
  \begin{center}
   \includegraphics[width=.6\linewidth,trim={0cm 0cm 0cm 0cm},clip]{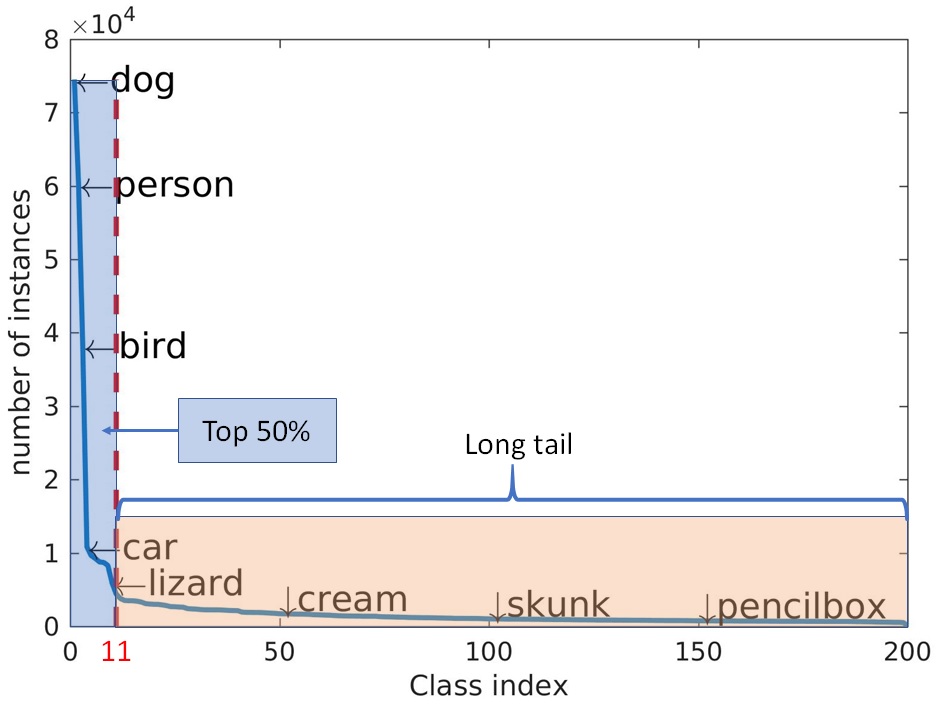}
  \end{center}
   \caption{Long-tail distribution of imageNet dataset}
\label{fig:longtail}
\end{figure}

We visualize the long-tail distribution of ILSVRC detection classes in Figure \ref{fig:longtail}. One can find that only 11 highly frequent classes (out of 200) cover top 50\% of the distribution. This distribution creates a significant impact on ZSD. To address this problem, in the second step of training, we augment the less frequent data to make a balance among similar seen classes for each unseen category. From the 10 million mini-batches used at the first stage of training, we create a set of over 2.8 million mini-batches for the second stage training. While creating this set, we make sure that every unseen class gets at least 10K similar (positive)  instances from classes whose meta-class category is common to that of unseen class. In doing so, for some unseen classes like `ray', we need to randomly augment data by repetition because the total instances of classes in the meta-class `non-mammal animal' are not more than 10K. In contrast, the unseen class like `tiger' has more than 10K similar instances in `mammal animal' meta-class. Therefore, we randomly pick 10K among those to balance the training set. After this, the rest of instances of 2.8 million mini-batches are chosen as the background.

    \begin{table}[!t]
  \begin{center}
  \scalebox{1.0}{
    \begin{tabular}{|c|c|c|c|}
    \hline
    mAP & Network  & w2v  & glo \\
    \hline\hline
    Baseline & R  & 31.0 &26.7 \\
    Our ($L_{cls}$) & R &\textbf{33.5}&\textbf{32.3}\\ 
    \hline 
	Baseline & V &30.3&27.9 \\ 
    Our ($L_{cls}$) & V &\textbf{30.4}&\textbf{28.4} \\ 
    \hline
    \end{tabular}}
  \end{center}
  \caption{ZSD on CUB using $\lambda = 1$. We refer V=VGG and R=ResNet}
  \label{tab:cub}
\end{table}

\section{ZSD on CUB} \label{sec:cub}
We evaluate the ZSD performance of the baseline and our proposed method based on single bounding box per image provided in CUB dataset \cite{CUB_2011}. Table \ref{tab:cub} describes the performance comparison between the baseline and our basic method. Our overall loss ($L_{cls}$) based method outperforms the baseline in the different network and semantic settings. Note that, we do not define any meta-class for the CUB classes. Therefore, we use $\lambda = 1$ for CUB related experiments.

\section{Further Analysis} \label{sec:further}

\begin{figure}[t]
  \begin{center}
\includegraphics[width=1\linewidth,trim={0cm .5cm 0cm 0cm},clip]{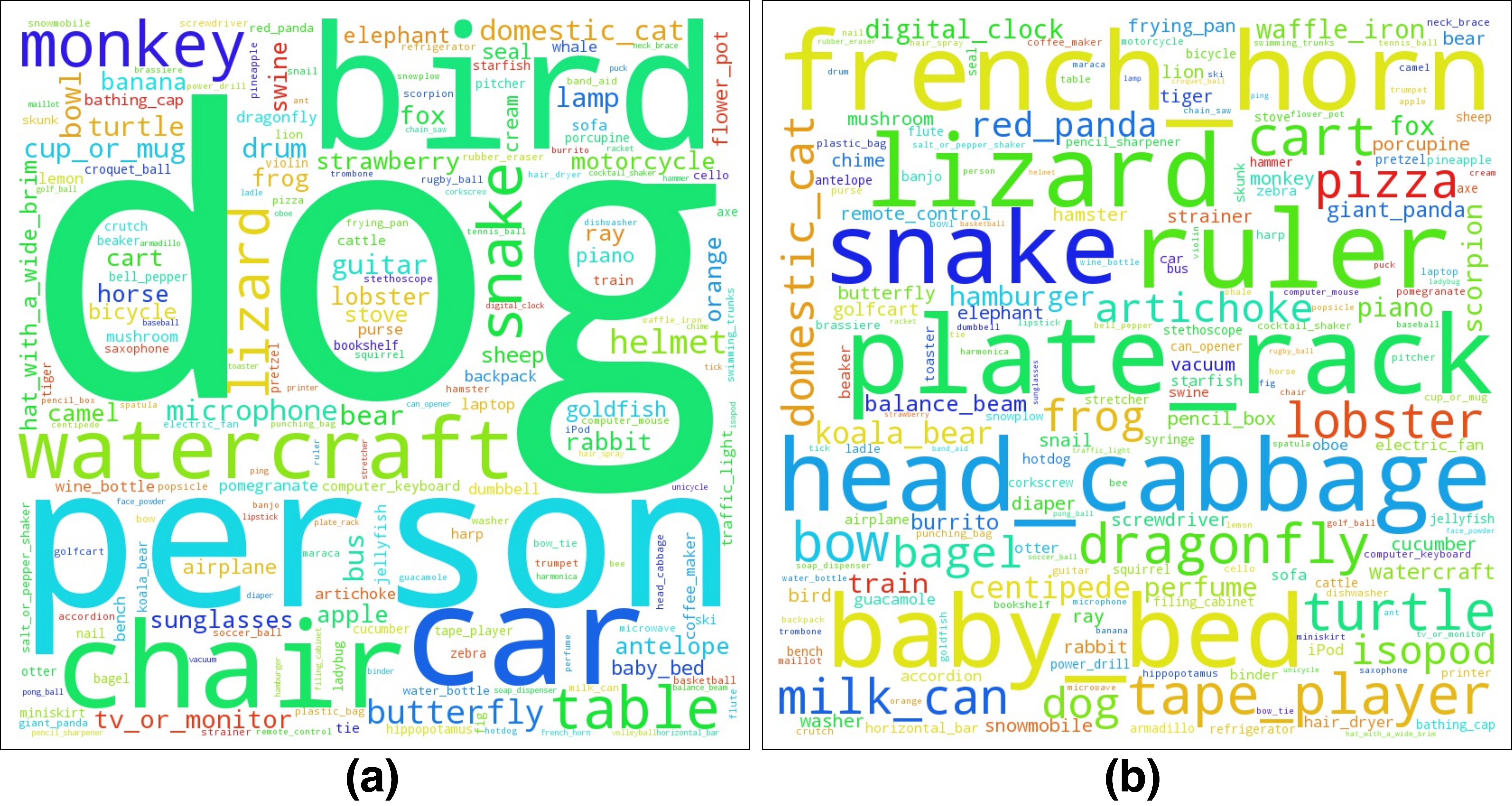}
  \end{center}
    \caption{Word cloud based on (a) number of object instance (b) Mean object size in pixel}
    \label{fig:wordcloud}
\end{figure}

\begin{figure*}[t]
  \begin{center}
\includegraphics[width=1\linewidth,trim={5.2cm 0cm 4cm 3cm},clip]{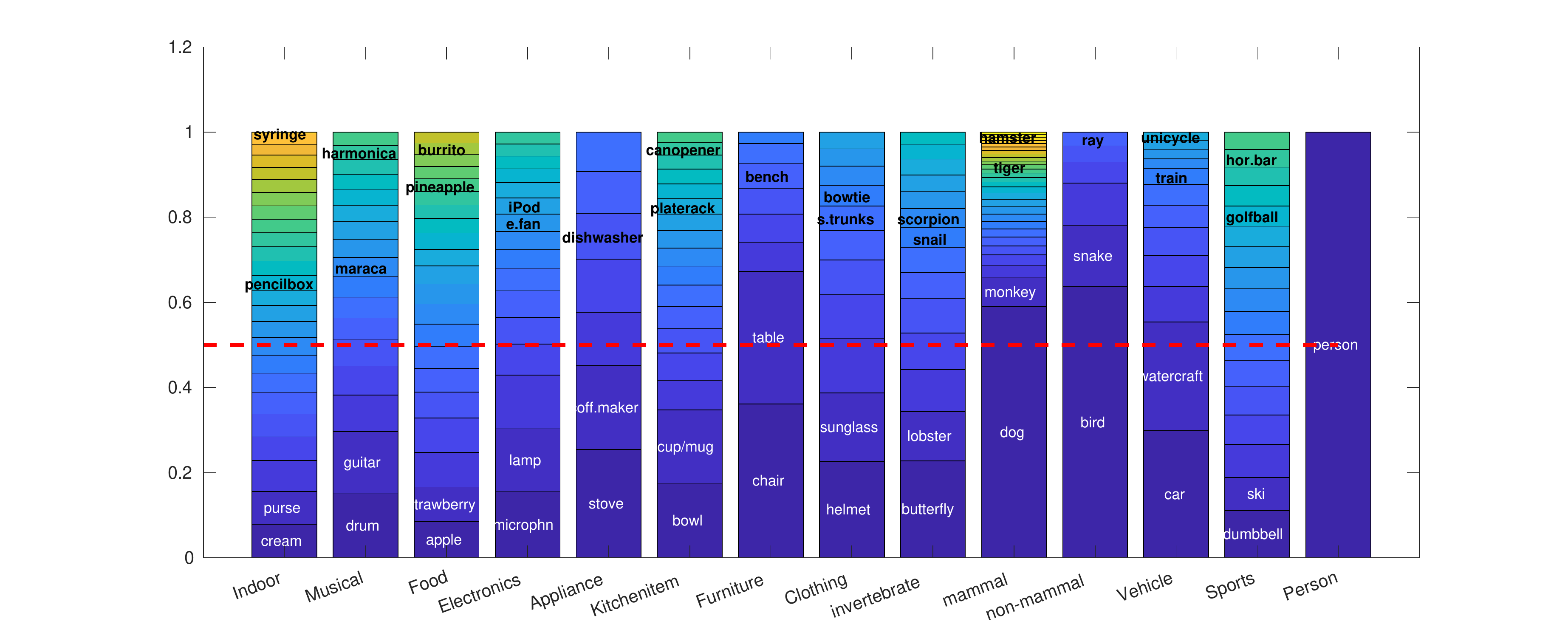}
  \end{center}
   \caption{Distribution of instances per classes within each meta class. Two most common (frequent) seen classes and unseen classes are marked in white and black color text respectively. Red dashed line indicates 50 percentile boundary. All unseen classes lie within the rarest half of the instance distribution.}
\label{fig:superclsbar}
\end{figure*}

\textbf{ZSD Challenges:} In general, detection is a harder task than recognition/tagging because of locating the bounding box at the same time. The strict requirement of not using any unseen class images during training of zero-shot setting is itself a tough condition for recognition/tagging task which gets intensified to a high degree for detection task. We have used ILSVRC-2017 detection dataset to evaluate some baseline performances of the proposed problem. This dataset has 200 classes including a total 478,807 object instances of different shapes/size and distribution (See Figure \ref{fig:wordcloud}). Within those, we define $\mathrm{M} = 14$ meta classes which contain one or more specific classes. Figure \ref{fig:superclsbar} describes the normalized number of instances per classes within meta class. Considering this challenging dataset, here we describe some other difficulties of the zero shot detection task:

\textit{Rarity:} ILSVRC dataset contains a long-tail distribution issue, i.e., many rare classes get less number of instances. It is apparent that an unseen class should be within the set of rare classes. To address this fact, we randomly choose unseen classes from each meta-class $z_j$ which lies in the rarest 50\% in the distribution. 
It affects the zero-shot version of the problem also.

\textit{Object size:} Some rare object classes like syringe, ladybug etc. usually have a small size. Smaller objects are difficult to detect as well as recognize.

\begin{figure*}[t]
  \begin{center}
\includegraphics[width=1\linewidth,trim={0cm 0cm 0cm 0cm},clip]{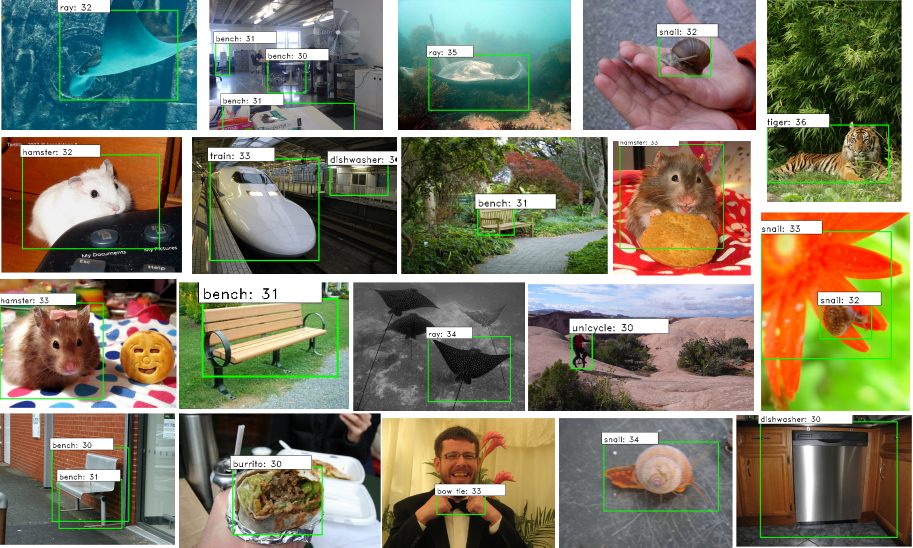}
  \end{center}
    \caption{Selected examples of ZSD of our ($L_{cls}$) with $\lambda=.6$ and R+w2v, using the prediction score threshold = 0.3.}
    \label{fig:output}
\end{figure*}

\begin{figure*}[!t]
  \begin{center}
\includegraphics[width=1\linewidth,trim={0cm 0cm 0cm 0cm},clip]{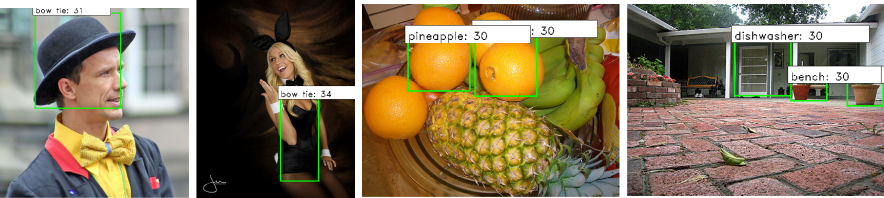}
  \end{center}
    \caption{Examples of incorrect detection but correct classification. The unseen class `bow-tie', `pineapple' and `bench' are incorrectly localized in these images.}
    \label{fig:output_neg}
\end{figure*}

\textit{High Diversity:} Every meta-class gets a different number of classes and there exists a high visual diversity in each meta-class images. Since, being in a same meta-class does not guarantee of the visual similarity, it is difficult to learn relationships for the unseen categories which are quite different from the seen categories in the same super-class. As an example, `tiger' has many similar classes compared to `ray'. The scarcity of similar class enables an inadequate description of the unseen class which eventually affect the zero shot detection performance.

\textit{Noise in semantic space:} We use unsupervised semantic embedding vectors word2vec/GloVe as the class description. Such embeddings are noisy in general as they are generated automatically from unannotated text mining. It also affects the zero-shot detection performance significantly.

\textbf{Seen vs. Unseen Class Performance:} The overall performance of ZSD is depended on the learning of seen classes. Therefore, the performance of seen detection can be an indication of how possibly ZSD works. To this end, we also study the detection performance on seen classes of ILSVRC validation dataset after the first step of faster-RCNN training (Table \ref{tab:seenresult}). It indicates the baseline performance of seen classes necessary to achieve the ZSD performance reported in the paper. The baseline method result is better than our proposed approaches. It is justifiable as both of our proposed methods can generate prediction for both seen and unseen class together which sacrifices the seen performance a bit to achieve distinction among all seen and unseen classes. The Table \ref{tab:seenresult} also compares the seen result with the unseen performance. One can find that performance of selected unseen classes is similar to that of seen classes for our ($L_{cls}$) method. It indicates a balanced generalization of ZSD in both seen and unseen classes.

\textbf{Learning without meta-class:} For some applications, the meta-class based supervision may not be available. In such case, one can define meta-class in an unsupervised manner by applying a clustering mechanism on original semantic embedding.

\textbf{ZSL vs ZSD loss:} Many traditional non-end-to-end trainable ZSR methods consider different aspects of regularization \cite{Morgado_2017_CVPR}, transductive setting \cite{Li_2017_CVPR}, metric learning \cite{bucher_ECCV_2016}, domain adaptation \cite{Kodirov_2015_ICCV} and class attribute association \cite{Al-Halah_2016_CVPR} etc. Similarly, the end-to-end trainable ZSR methods \cite{Zhang_2017_CVPR,Ba_CVPR_2015} employ different non-linearity in feature and semantic pipeline. But, those traditional loss formulations need to be redesigned in ZSD case to be compatible for both classification and box detection losses.

 \begin{table}[!t]
  \begin{center}
  \scalebox{0.9}{
    \begin{tabular}{|c|c|c|c|c|}
    \hline
    mAP & Step 1 & Baseline  & Ours ($L'_{mm}$)  & Our ($L_{cls}$)  \\
    \hline\hline
	Seen & \textbf{33.7} &33.4&27.7&26.1 \\ \hline
    Unseen (all) & - &12.7&15.0&\textbf{16.4} \\ \hline
    Unseen (setected) & - &18.6&22.7&\textbf{27.4} \\ \hline
    \end{tabular}}
  \end{center}
  \caption{Comparison of seen and unseen class performance using ResNet as convolution layers. word2vec is used for baseline, our ($L'_{mm}$) and our ($L_{cls}$). Best performance in each row are shown as bold. We refer Unseen (all): mAP of all unseen classes, Unseen (selected): mAP of selected classes for which visually similar classes are present.}
  \label{tab:seenresult}
\end{table}

\section{Qualitative results} \label{sec:qualitative}

We provide more examples of ZSD in Fig. \ref{fig:output}. One can find that the prediction score threshold is lower (0.3 used in the examples) than the value (greater than 0.5) used in traditional object detection like faster-RCNN \cite{Faster_RCNN_2017}. It indicates that the prediction of ZSD has less confidence than that of traditionally seen detection. As zero-shot method does not observe any training instances of unseen classes during the whole learning process, the confidence of prediction cannot be as strong as the seen counterpart. Moreover, a ZSD method needs to correspond visual features with semantic word vectors which are noisy in general. It degrades the overall confidence for ZSD.

In the last layer of the box regression branch, our method does not have specified bounding boxes for un-seen classes. Instead, bounding box corresponding to a closely related seen class that has the maximum score is used for un-seen localization. Therefore, a correct unseen class prediction sometimes cannot get very accurate localizations as illustrated in Fig. \ref{fig:output_neg}.

\end{document}